\pdfoutput=1
\documentclass[11pt]{article}

\usepackage[]{acl}

\usepackage{times}
\usepackage{latexsym}

\usepackage[T1]{fontenc}
\usepackage[utf8]{inputenc}

\usepackage{microtype}

\usepackage{pifont}

\usepackage{graphicx}
\usepackage{cleveref}
\usepackage{pgfplots}
\usepackage{caption}
\usepackage{subcaption}
\usepackage{bm}

\title{What do Models Learn From Training on More Than Text? \\Measuring Visual Commonsense Knowledge}

\author{Lovisa Hagström \\
  Chalmers University of Technology \\
  \texttt{lovhag@chalmers.se} \\\And
  Richard Johansson \\
  University of Gothenburg \\
  \texttt{richard.johansson@gu.se} \\}

\begin{document}
\maketitle
\begin{abstract}
There are limitations in learning language from text alone. Therefore, recent focus has been on developing multimodal models. However, few benchmarks exist that can measure what language models learn about language from multimodal training. We hypothesize that training on a visual modality should improve on the visual commonsense knowledge in language models. Therefore, we introduce two evaluation tasks for measuring visual commonsense knowledge in language models\footnote{Code publicly available at: \url{github.com/lovhag/measure-visual-commonsense-knowledge}} and use them to evaluate different multimodal models and unimodal baselines. Primarily, we find that the visual commonsense knowledge is not significantly different between the multimodal models and unimodal baseline models trained on visual text data.
\end{abstract}

\section{Introduction}
% Light on problem with models that only learn language from language due to nature of language (Bender) and reporting bias
% Claims that slow LM learning may come from slow learning of commonsense knowledge (When do you need billions of training data)
% Recently, many models that combine vision and text proposed
% Mostly evaluated on vision+text tasks, there are few evaluation tasks that measure the textual properties of the model.
% Does a multimodal model learn visual commonsense knowledge?

%Grounding has been presented as the solution to current issues with LMs suffering 

Language models (LMs) trained on large amounts of textual data have shown great performance on several textual tasks \citep{devlin2018bert,gpt3}. However, recent work has illuminated limitations with text-only training of LMs. These limitations arise from a lack of meaning \citep{bender-koller-2020-climbing} and experience \citep{bisk-etal-2020-experience}, together with the problem of reporting bias \citep{gordon2013reporting}. Multimodal training has been identified as one way to create models that do not suffer from the aforementioned limitations \citep{paik-etal-2021-world,huang2021makes}. While several multimodal models have been developed \citep{tan-bansal-2019-lxmert,li2019visualbert,li2020oscar}, few evaluation methods exist that can tell us whether multimodal training mitigates text-only training limits.
\begin{figure}[t]
    \centering
    \includegraphics[scale=0.18]{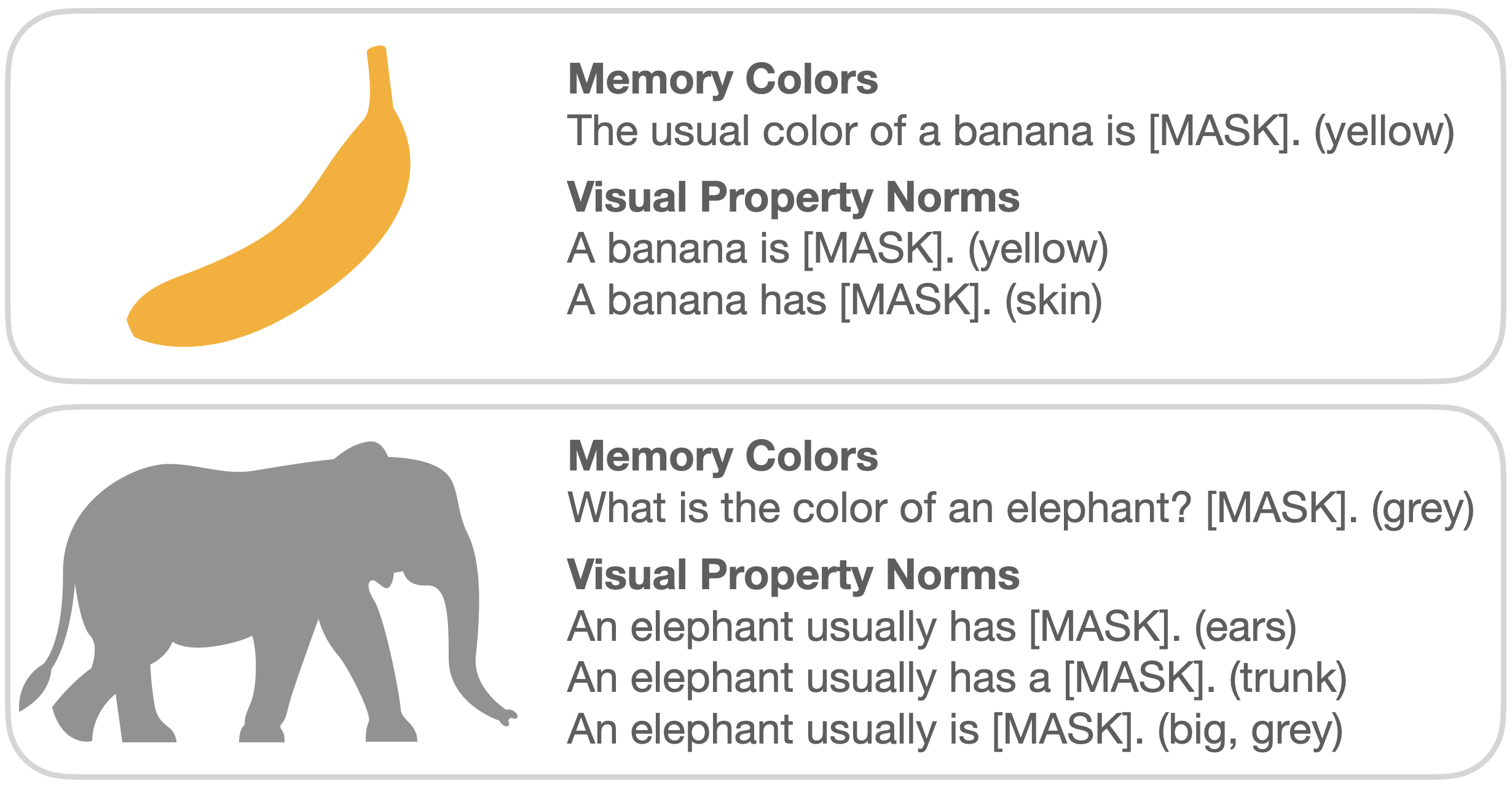}
    \caption{We introduce the two evaluation tasks Memory Colors and Visual Property Norms for measuring visual commonsense knowledge in a LM.}
    \label{fig:overview}
\end{figure}

If we wish to successfully create multimodal LMs that learn from more than text, we need a way to evaluate them for what we expect them to have learned from their multimodal training. %We also need to make sure that the models can leverage this knowledge in their language usage. %In essence, we need a way to more closely measure how \emph{grounded} a LM is.

One hypothesis is that multimodal training should aid LMs in learning commonsense knowledge \citep{zhang-etal-2021-need}. There are several text-only evaluation tasks that aim to measure the commonsense knowledge in LMs \citep{zellers-etal-2019-hellaswag,Zhou_Zhang_Cui_Huang_2020}, but none of them focus explicitly on the commonsense knowledge learned through training on more than text. 

In this work, we focus on models trained on images and text, denoted \emph{vision-and-language models}. We reason that if there is any additional information to be learnt from a visual modality it should firstly be basic visual commonsense knowledge. That is, visual conceptual knowledge that is viewed as commonsense by humans, and thus not attainable from text alone due to reporting bias. %why basic?

%A simple way to measure what a model has learned from multimodal training is to measure what commonsense knowledge it has learned from its additional modality, where we expect different types of commonsense knowledge to be learned from different types of modalities. 

We propose a simple method for measuring the visual commonsense knowledge of a model using two zero-shot masked language text-only tasks, depicted in \Cref{fig:overview}. The first task is the Memory Colors evaluation task \citep{norlund2021} and the second we create based on the visual features in the Centre for Speech, Language and the Brain (CSLB) concept property norms dataset \citep{devereux2014centre}. We refer to the latter task as the Visual Property Norms evaluation task. We complement our work with the results of four vision-and-language models and four baselines on these two tasks.

\section{Evaluation Tasks}

Our aim is to evaluate models for visual commonsense knowledge. To do this we make use of the existing Memory Colors evaluation task described in \cref{sec:memory-colors}, and introduce a new evaluation task, Visual Property Norms in \cref{sec:visual-property-norms}. Memory Colors is smaller than Visual Property Norms and specifically focuses on visual information related to the color of different concepts, so it is potentially easier. We include both tasks to get a performance curve over increasing difficulty.

Common for both tasks is that they contain queries in English relating to visual properties of tangible concepts and that these queries are based on the knowledge of multiple human participants. Therefore, the tasks can be considered to evaluate a basic aspect of visual commonsense knowledge.% and model scores on these tasks should indicate how much commonsense knowledge a model has.

Also common for both tasks is that they use textual templates containing a \texttt{[MASK]} token to be predicted by a model in a cloze-style fashion, similarly to the method used by \citet{kassner-schutze-2020-negated} and \citet{petroni-etal-2019-language}. The advantages with querying the models in this fashion is that most LMs\footnote{Excluding autoregressive LMs.} already have been exposed to this type of query format, including most multimodal models. We can then evaluate any model in a masked language modelling fashion on these tasks without additional training or having to make model-specific adaptations, enabling easy evaluation for researchers who wish to use these evaluation tasks.

This form of cloze-style evaluation is also referred to as \emph{prompt-based retrieval}. The reliability of this method has recently been questioned by \citet{jiang2020know} and \citet{cao-etal-2021-knowledgeable} due to the query format sensitivity of LMs. To alleviate this issue, we evaluate the models using several different prompts for each of the two tasks.

%This is necessary to get a robust estimate of the performance of the evaluated LMs, since LMs are sensitive to query format \citep{jiang2020know}.

\subsection{Memory Colors} \label{sec:memory-colors}

The Memory Colors evaluation task is a text-only zero-shot cloze test in English that evaluates a model for its knowledge of memory colors. It queries a model for the color of 109 typical objects using 13 different query templates. The task has been created with the help of 11 human participants, so to some extent it encodes human visual commonsense knowledge limited to colors. Some examples of queries can be seen in \Cref{fig:overview}.

We use the same evaluation metric as specified by \citet{norlund2021}, i.e. the accuracy score after masking the model output for the 11 possible colors black, blue, brown, green, grey, orange, pink, purple, red, white and yellow.

\subsection{Visual Property Norms}\label{sec:visual-property-norms}

We also introduce a new cloze task in English to evaluate for visual commonsense knowledge, denoted Visual Property Norms. It is the largest query-based pure-language evaluation task capable of evaluating LMs for visual commonsense knowledge, containing 6,541 visual conceptual features produced by human participants. 

We base it on the CSLB concept property norms dataset \citep{devereux2014centre} that contains the conceptual knowledge of 30 human participants for each of 541 concrete objects, with 123 participants in total. This knowledge is represented as a set of features per object, for which each feature is specified with a production frequency (PF). The PF describes how many of 30 participants produced that feature, so a feature with a high PF can be considered to be more apparent to the participants, since more came to think of it. All features are also categorized as either \emph{encyclopaedic}, \emph{functional}, \emph{other perceptual}, \emph{taxonomic} or \emph{visual perceptual}. \Cref{tab:normdata_examples} contains some examples of visual perceptual features in the dataset. 
%
%Only features that have been produced by at least two participants are included in the dataset, so to some extent it can be viewed as containing conceptual commonsense knowledge.
%
\begin{table}[h]
    \centering
    \begin{tabular}{l|l|l|r}
        Concept & Relation & Feature & PF \\
        \hline
        Cherry & has a & stalk & 17 \\
        Fern & is & green & 29 \\
        Hair & is & thin & 22 \\
        Plum & has & flesh & 9 \\
    \end{tabular}
    \caption{Some concepts and their visual perceptual features in the concept property norms dataset.}
    \label{tab:normdata_examples}
\end{table}

We create our evaluation task from the concept property norms dataset in a set of steps. Firstly, since our goal is to measure visual commonsense knowledge, we only make use of the \emph{visual perceptual} features. Since we wish to perform cloze tests through masked language modelling, only feature alternatives describable by one wordpiece from the BERT base uncased tokenizer are included.

Furthermore, we only include the four most common feature relations in the task. These are \emph{has}, \emph{has a}, \emph{made of} and \emph{is}. We then part the data into five different segments based on production frequency. This is done by thresholding the features for each concept such that only features with a PF above the set threshold for a certain data segment are included as gold labels in that segment. %Features with higher PFs can be considered to be more apparent. %This means that the segments with higher PF thresholds contain gold labels that are subsets of the labels in the segments with lower PF thresholds. 
%To some extent, the segments with higher PF thresholds can be viewed as containing ``more common'' visual commonsense knowledge. 
The segments and their PF thresholds are listed in the appendix.

Lastly, we create queries from the concepts in each data segment using 8 different query templates, seen in the appendix. Some examples of Visual Property Norms queries can be seen in \Cref{fig:overview}.

Similarly to \citet{Weir-et-al:2020} we use the mean average precision (mAP) as our evaluation metric, since there may be multiple correct answers for each query in our evaluation data. We calculate this score for each concept and relation, per query template and production frequency segment. We then get a final score for each production frequency segment by taking the average score over all query templates and concepts per segment.
%\begin{equation}
%    \mathrm{s} = %\frac{1}{n_{\mathrm{q}}}\sum_i^{n_{\mathrm{q}}} %\frac{1}{n_{\mathrm{c}}}\sum_j^{n_{\mathrm{c}}} %\frac{1}{n_{\mathrm{r,j}}}\sum_k^{n_{\mathrm{r,j}}} \sum_t^{\mathrm{|vocab|}} P_{i,j,k}(t)\Delta r_{i,j,k}(t)
%    \label{eq:score}
%\end{equation}
%Here, $n_{\mathrm{q}}$, $n_{\mathrm{c}}$ and $n_{\mathrm{r,j}}$ respectively denote the number of query templates, concepts and relations per concept. $P_{i,j,k}(t)$ is the precision of the model for query template $i$, concept $j$ and relation $k$ after including the top $t$ token predictions of the model. $\Delta r_{i,j,k}(t)$ similarly measures the change in recall from including the first $t-1$ top predictions to also including the $t$th top prediction of the model. 
This metric is measured over a vocabulary that has been masked to only contain the 614 possible answer alternatives in the Visual Property Norms evaluation data. %Thus, $\mathrm{|vocab|}=614$ in \cref{eq:score}.

%The difference between Memory Colors and Visual Property Norms lies in that the latter contains more general queries with visual perceptual answer alternatives that aren't only limited to color. Visual Property Norms is also a larger evaluation task with more queries to base the model evaluation on.

%To obtain a deeper insight into the model performance, we also report the masked scores per query template.

%To easier compare performance on the property norms evaluation task with performance on the memory colors task, we report the masked score on a subset of the property norms data that only contains queries for which the gold labels are one or several colors from the Memory Colors task. 

%\section{Experimental setup}

\begin{figure*}[t]
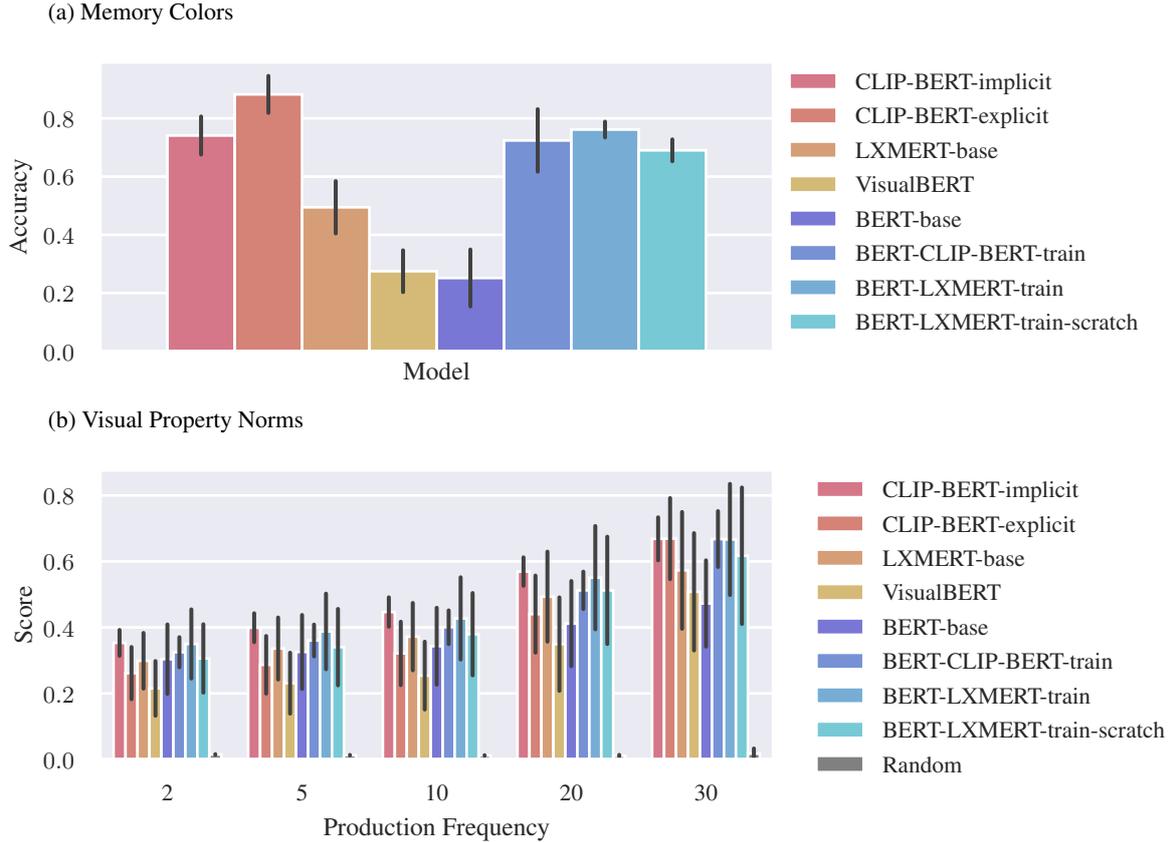

    \begin{center}
    \begin{subfigure}[b]{0.9\textwidth}
    \caption{Memory Colors}
    \label{fig:memory-colors-results}
    \hspace{-0.8cm}\input{images/memory_colors_results.pgf}
    \end{subfigure}
    \begin{subfigure}[b]{0.9\textwidth}
    \caption{Visual Property Norms}
    \label{fig:normdata-results}
    \hspace{-0.8cm}\input{images/all.pgf}
    \end{subfigure}
    \end{center}
    \caption{The model accuracy on Memory Colors and model scores on Visual Property Norms per production frequency segment. The multimodal model results are depicted with warmer colors, and the unimodal model results are depicted in cooler colors. The error bars indicate the standard deviation of the model performance over the different query templates. The score has been calculated by masking the vocabulary of the models to only contain the possible answers of the task.}
    \label{fig:results}
\end{figure*}

\section{Models}

We evaluate four multimodal pre-trained models for their visual commonsense knowledge. These are CLIP-BERT both with and without imagination\footnote{The explicit version has the ability to ``imagine'' visual features when queried with text.}\citep{norlund2021}, a LXMERT base uncased \citep{tan-bansal-2019-lxmert} and VisualBERT %pre-trained on MS COCO and VQA 
\citep{li2019visualbert}. We also evaluate four unimodal baseline models. These are a BERT base uncased pre-trained on English Wikipedia and BookCorpus, a BERT base uncased further trained on the pure-text part of the CLIP-BERT training data (BERT-CLIP-BERT-train) and two BERT base uncased models trained on the pure-text part of the LXMERT training data, one from scratch and one initialized from pre-trained BERT weights (BERT-LXMERT-train-scratch and BERT-LXMERT-train).

All models are to some extent based on the BERT base architecture and consequently share the same vocabulary and tokenizer. They are also of similar sizes with $\sim 110$M trainable weights, the exception being LXMERT with $\sim 230$M trainable weights. %The main differentiating aspects of the models are their pre-training data, pre-training objectives and different backbones for the visual features. 
Additional information about the models can be found in the appendix. 

\vspace{-0.2cm}
\paragraph{Adapting the models for pure-text queries} The majority of current multimodal models have not been developed to be queried only with text. In this case, both CLIP-BERT and VisualBERT should work well with only removing their visual features input, since they are single-stream models. However, LXMERT is a dual-stream model that requires a visual feature input. %A way to handle the removal of visual information for LXMERT in previous work has been to supply the model with an average feature vector of the current evaluation dataset \citep{frank-etal-2021-vision}. 
We handle the removal of visual information by simply removing the visual processing chain in LXMERT, making the language input the only input given to the Cross-Modality Encoder in the model. This would not work if we still wanted to use the model in a multimodal fashion, but we can make this adaption since we are only interested in querying the model for visual commonsense knowledge via language.

%\subsection{Evaluation}

%We take each pre-trained model as listed in \Cref{tab:model-pretraining-data} and evaluate it through pure-text MLM on both Memory Colors and Visual Property Norms. We perform no fine-tuning of the models on the evaluation tasks. For Memory Colors we report the accuracy and for Visual Property Norms we report the mAP score for each production frequency segment. %We report the score both with and without masking the output vocabulary for the 614 possible label alternatives.

\section{Results}

%Add some concepts and predictions. Motivate masking with prediction examples for ``is''. Reason about CLIP-BERT explicit worse performance?

%\begin{table}[h]
%    \centering
%    \begin{tabular}{l|c}
%    \hline
%    Model & Accuracy (k=1) \\
%    \hline
%    Human baseline & $\bm{0.937} \pm 0.051$ %\\
%    Majority baseline & $0.229 \pm 0.000$ %\\
%    BERT     & $0.252 \pm 0.102$ \\
%    \hline
%    CLIP-BERT-implicit & $0.741 \pm %0.068$\\
%    CLIP-BERT-explicit & $\bm{0.882} \pm %0.066$\\
%    BERT-CLIP-BERT-train & $0.724 \pm %0.112$ \\
%    \hline
%    LXMERT & $0.495 \pm 0.094$\\
%    BERT-LXMERT-train & $\bm{0.761} \pm %0.028$ \\
%    BERT-LXMERT-train-scratch & $0.690 \pm %0.039$ \\ 
%    \hline
%    VisualBERT & $0.275 \pm 0.075$\\
%    \end{tabular}
%    \caption{Model results on Memory %Colors. Standard deviation is reported %over query templates. The top three %model results are marked in bold.}
%    \label{tab:memory-colors-results}
%\end{table}

The results of the models on our two evaluation tasks can be seen in \Cref{fig:results}. We format the analysis of the results around a set of questions.

\textbf{Do the multimodal models display more memory colors knowledge?}
  The multimodal CLIP-BERT-explicit model has the best performance on this task. So to some extent, yes. But it is worth noting that the unimodal BERT model trained on LXMERT training data is second best on the task, outperforming both LXMERT and VisualBERT, indicating a small multimodal advantage.
 
 %\paragraph{Does model ``imagination'' improve Memory Colors performance?}
 %CLIP-BERT-explicit utilizes imagination and has the best performance on this task, also compared to its counterpart CLIP-BERT-implicit that doesn't use imagination. So imagination seems to help model performance on Memory Colors.
 
 %\vspace{-0.2cm}
 \textbf{Is performance on Memory Colors indicative of performance on Visual Property Norms?}
 The ranking visible in \Cref{fig:memory-colors-results} does not entirely differ from that in \Cref{fig:normdata-results}. 
 The main exception being CLIP-BERT-explicit, which has the best performance on Memory Colors, but is outperformed by most other models on Visual Property Norms. We perform a closer analysis of how these results compare by extracting Visual Property Norm results for colors in the appendix. 
 
%\vspace{-0.2cm}
\textbf{Do the models perform better when evaluated on more apparent concept features?}
We can observe how the model performance unanimously increases with increased production frequency threshold in \Cref{fig:normdata-results}. Thus, it appears as though the models agree more with concept features that can be regarded as more apparent. 

%\vspace{-0.2cm}
\textbf{Do the multimodal models contain more visual commonsense knowledge?}
The results in \Cref{fig:normdata-results} do not really indicate clear advantage of either unimodal or multimodal models. The multimodal model CLIP-BERT-implicit may generally have the best performance on the task, but the unimodal models trained on visual text data do not differ much in performance. For example, the unimodal BERT-LXMERT-train performs almost on par with CLIP-BERT-implicit.

This conclusion is similar to that of \citet{yun-etal-2021-vision-language}, who also compared vision-and-language models to text-only models trained on captions. They found that the models have similar performance with respect to their internal linguistic representations for general tasks.

These results do not mean that the idea of having models learn language from more than text has failed. They do however indicate that there is more work to be done on developing models that use multimodal pretraining to improve on their natural language understanding. %Example of such work is to create models designed to work well both in a multimodal domain, and in a text-only domain.

However, we cannot exclude the possibility in our work that the multimodal models suffer in performance due to a lack of visual feature input. %, since the models have been fine-tuned to process both text and image. 
Future work investigating this would be valuable.

%\vspace{-0.2cm}
\textbf{Are the models sensitive to how they are queried?}
Prevalent for all models is that their performance varies greatly with how they are queried.  BERT-LXMERT-train may have the best performance on Visual Property Norms if queried differently. We evaluate the model performances depending on query template in the appendix. This highlights the importance of querying the models with different prompts, since the models may perform dissimilarily depending on prompt due to the degree of prompt-dataset fitness, as reported by \citet{cao-etal-2021-knowledgeable}.

%\vspace{-0.2cm}
\textbf{Does fine-tuning on visual language develop visual commonsense knowledge?}
In both \Cref{fig:memory-colors-results,fig:normdata-results} it is visible that unimodal model performance greatly improves with fine-tuning on visual text corpora. Potential explanations for this are that the models become more attuned to the task with fine-tuning, or that corpora from VQA and image captioning do not suffer as much from reporting bias compared to more common corpora. Thus, text that has been curated to explicitly contain visual information may suffice as a replacement for images.

\section{Related Work}

\citet{Weir-et-al:2020} also use the CSLB concept property norms to probe LMs for commonsense knowledge. Our work differs from theirs in that we focus on visual commonsense knowledge and evaluate several multimodal models for whether their multimodal training grants them additional visual commonsense knowledge.

\citet{norlund2021} also query a multimodal model for visual commonsense knowledge but with a focus on memory colors. \citet{paik-etal-2021-world} present similar work but with more focus on probing and reporting bias. In our work, we include general visual commonsense knowledge concepts and evaluate several multimodal models.

Additionally, \citet{iki-aizawa-2021-effect} evaluate several vision-and-language models on GLUE, to investigate the effect of an additional visual modality on the general linguistic capabilities of a model. Our work differs in that we evaluate the models specifically for visual commonsense knowledge.  %However, our work agrees with theirs in that vision-language-models do not improve

Other tasks that have been developed to evaluate the performance of vision-and-language models are Visual Question Answering (VQA) tasks and Visual Commonsense Reasning (VCR) tasks \citep{balanced_vqa_v2, hudson2018gqa,zellers2019vcr}. Our work differs from these in that we evaluate for visual knowledge in models without conditioning on an image, to investigate whether the linguistic capabilities of a model improve from training on more than text. In the aforementioned tasks, the text prompts are always conditioned on an image provided with the prompt, obstructing equal comparisons with text-only models.

%Other color article?

\section{Limitations}
Our work is limited to a subset of vision-and-language models, so the results found may not translate to all such model types. Also, since our evaluation utilizes prompt-based retrieval, its measurement accuracy depends on how well this method works for LMs.
Additionally, as previously mentioned, we do not investigate how well the multimodal models adapt to a unimodal input. Thus, our results depend on whether the models were functioning adequately with our method of adapting them to a unimodal input.

\section{Ethical Considerations}
Our work should not have any direct ethical implications, since we mainly introduce evaluation tasks and evaluate different models on them. We do however investigate visual conceptual perceptions based on data from a potentially small group of people whose world-view may be culturally different from that of other individuals. This means that we may encourage knowledge that benefits some people more than others.
Similar issues are discussed by \newcite{liu2021visually}.
Our investigation is limited to English-language models and datasets, limiting the generality of our conclusions.

\section{Conclusions}
We introduce new evaluation methods for measuring the visual commonsense knowledge in LMs and evaluate a number of multimodal LMs on these benchmarks. We find that there are no significant differences in performance between models trained on pure text and models trained on images and text. Most prominently, we find that a unimodal LM trained on image captions and VQA queries can attain a visual commonsense knowledge on par with that of a multimodal model.

%The questions are then; what language aspects can we expect language models trained on more than text to improve in? And how can we measure it? 

We also confirm the results by \citet{jiang2020know} and \citet{cao-etal-2021-knowledgeable}, that LMs are sensitive to query format even when querying for commonsense knowledge. This casts some doubts on what is really measured in a model for a cloze task and whether we can reason about LMs as having knowledge. An interesting future step would be to investigate this further and see if it would be more applicable to use e.g. probing or some other evaluation method.%some communicative task rather than cloze-style queries, or knowledge bases combined with LMs. %towards quantifying how ``grounded'' a model is would be to find a more accurate evaluation method. 

Nonetheless, this is a first step towards measuring the visual commonsense knowledge in multimodal as well as unimodal LMs. We hope that the evaluation tasks introduced here may aid other researchers in their aim to create models that learn language from more than text.

%Evaluate through solving some task

%Knowledge acquisition speed (from SCRATCH) interesting avenue

\section*{Acknowledgements}

We thank the anonymous reviewers for their valuable feedback and knowledge sharing.

This work was partially supported by the Wallenberg AI, Autonomous Systems and Software Program (WASP) funded by the Knut and Alice Wallenberg Foundation.

Additionally, the computations were enabled by resources provided by the Swedish National Infrastructure for Computing (SNIC), partially funded by the Swedish Research Council through grant agreement no. 2018-05973.

% Entries for the entire Anthology, followed by custom entries
\bibliography{anthology,custom}
\bibliographystyle{acl_natbib}

\appendix

\section{Additional model information}
\label{sec:model-appendix}
Additional information about the models used in our work and their training datasets can be found in \Cref{tab:model-pretraining-data,tab:datasets}. We can observe that VisualBERT has been trained on a data amount that is quite small compared to those of CLIP-BERT and LXMERT.

It is also worth noting on the different backbones of the models. CLIP-BERT is a single-stream multimodal model with a CLIP backbone for visual processing. LXMERT is a dual-stream multimodal model with a Faster R-CNN detector backbone. While VisualBERT is a single-stream model that also utilizes Faster R-CNN detector backbone. Since CLIP has been trained on the immense WIT dataset, the backbone data sizes differ greatly between CLIP-BERT and the other multimodal models.

\begin{table*}
    \centering
    \begin{tabular}{l|l l l|l|l}
    \hline
    Model & Text & Visual text & Images+Text & Backbone & Training objectives \\
    \hline
    BERT & 80M &  &   & & MLM, NSP\\
    \hline
    CLIP-BERT-implicit & 80M &  & 4.7M & 400M & MLM \\
    CLIP-BERT-explicit & 80M &  & 4.7M & 400M & MLM \\
    BERT-CLIP-BERT-train & 80M & 4.7M &  & & MLM  \\
    \hline
    LXMERT &  &  & 9.2M & 0.1M & MLM, RFR, DLC, \\
    & & & & & ITM, IQA\\ 
    BERT-LXMERT-train & 80M & 9.0M & & & MLM \\
    BERT-LXMERT-train-scratch &  & 9.0M & & & MLM \\
    \hline
    VisualBERT & 80M &  & 1.7M & 0.1M & MLM, ITM \\
    \end{tabular}
    \caption{An overview of the pre-trained models, the sizes of their training datasets and their pre-training objectives. The sizes are measured in number of training samples. The backbone column indicates the training data sizes for the image processing backbones of the models. For the training objectives, ITM refers to Image-Text Matching, RFR to RoI-Feature Regression, DLC to Detected Label Classification, MVM to Masked Visual Modeling and IQA to image QA.}
    \label{tab:model-pretraining-data}
\end{table*}

\begin{table*}[]
    \centering
    \begin{tabular}{l|l|l|l|l}
        Dataset & Data sources & \# of text & \# of images \\
        \hline
        CLIP-BERT V+L & MS COCO, SBU Captions, VG-QA, CC & 4.72M & 2.91M \\
        LXMERT V+L & MS COCO, VG, VQA, GQA, VG-QA & 9.18M & 0.18M \\
        VisualBERT V+L & MS COCO, VQA & 1.27M & 0.12M
    \end{tabular}
    \caption{The vision-language datasets on which the multimodal models originally were trained. More information about the datasets can be found in the articles that introduced the models.}
    \label{tab:datasets}
\end{table*}

\section{Additional information on Visual Property Norms}\label{sec:normdata-info}
Information about the different segments and number of entries per segment in the Visual Property Norms can be seen in \Cref{tab:pf-segments}.

\begin{table}
    \centering
    \begin{tabular}{r|r|r r r r}
        PF & entries & \emph{has} & \emph{has a} & \emph{made of} & \emph{is}\\
        \hline
        2 & 6,541 & 1,675 & 1,190 & 1,176 & 2,500 \\
        5 & 3641 & 1,016 & 642 & 760 & 1,223\\
        10 & 2001 & 583 & 347 & 509 & 562 \\
        20 & 613 & 169 & 88 & 209 & 147\\
        30 & 27 & 5 & 2 & 10 & 10  
    \end{tabular}
    \caption{The data segments segmented based on production frequencies together with their number of entries. The entries are calculated as the number of feature-concept-label entries, where there can be several features belonging to the same feature and concept. The PF column indicates the production frequency threshold for each segment, all features with a production frequency higher or equal to this threshold are included in the segment. We also list the number of labels per feature relation type.}
    \label{tab:pf-segments}
\end{table}

\section{Additional results on Visual Property Norms}
\label{sec:normdata-appendix}

Additional model results on the Visual Property Norms can be found here. 

\begin{figure*}
    \centering
    \input{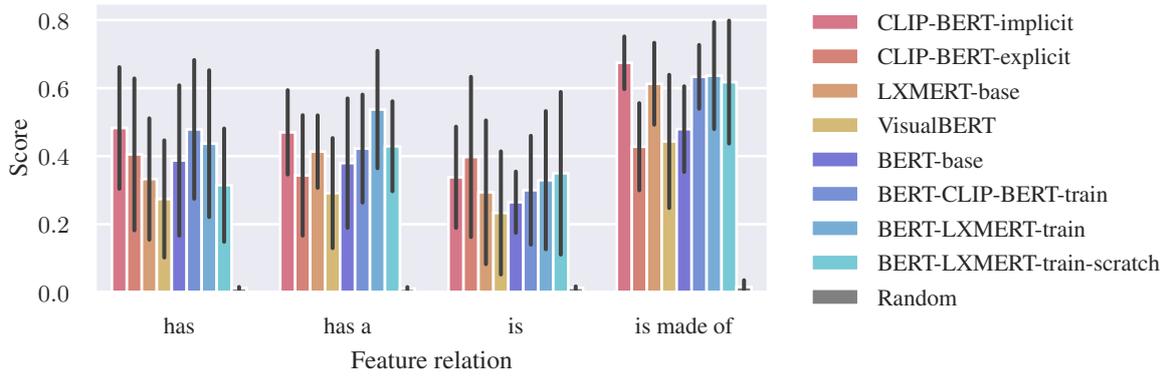}
    \caption{The model scores on Visual Property Norms per feature relation. The error bars indicate the standard deviation of the model performance over the different query templates. The score has been calculated by masking the vocabulary of the models to only contain the possible answers of the task.}
    \label{fig:score-per-relation}
\end{figure*}

\Cref{fig:score-per-relation} indicates model performance per feature relation across the production frequency segments. We can observe how the models show the best performance for the \emph{is made of} relation, which arguably can be associated more with visual perceptual properties. 

\begin{figure*}
    \centering
    \input{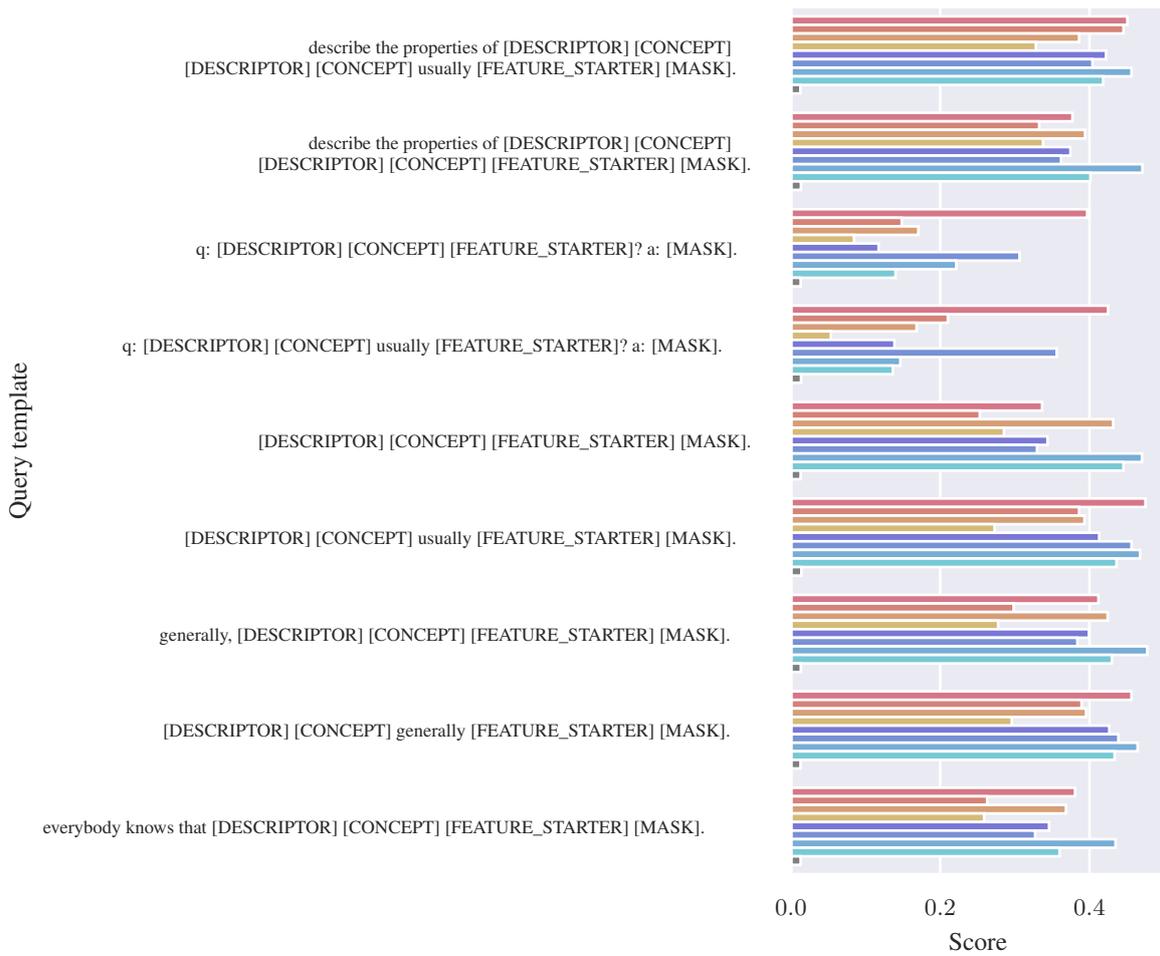}
    \caption{The score for each model on Visual Property Norms per query template. The score has been calculated by masking the vocabulary of the models to only contain the possible answers of the task.}
    \label{fig:score-per-query}
\end{figure*}

\Cref{fig:score-per-query} shows model score per query template across all production frequency segments, indicating that CLIP-BERT-implicit benefits from being more robust to different query templates. Additionally, these results indicate that BERT-LXMERT-train would have the best overall score on Visual Property Norms if the queries containing ``q: a'' were to be removed. 

\begin{figure*}
    \centering
    \input{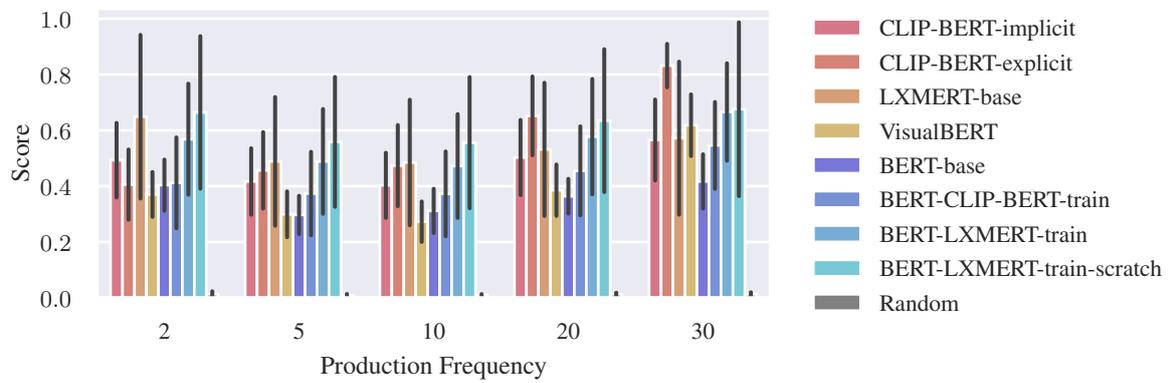}
    \caption{The score for each model per production frequency segment on Visual Property Norms that has been filtered to only contain samples for which the correct answer is one or more out of 11 possible colors. The score has been calculated by masking the vocabulary of the models to only contain the possible answers of the task.}
    \label{fig:score-normdata-colors}
\end{figure*}

Lastly, \Cref{fig:score-normdata-colors} contains the results of the models on the color part of Visual Property Norms which has been filtered to only contain queries with gold labels describing colors. Here, we see some indications of a better performance of CLIP-BERT-explicit for colors. Potentially, the imagination capacity of this model is more helpful for queries with answers relating to more basic visual properties, such as color.

\end{document}